# PARSING USING LINEARLY ORDERED PHONOLOGICAL RULES

Michael Maxwell

Summer Institute of Linguistics

7809 Radin Road

Waxhaw, NC 28173 USA

Internet: Mike.Maxwell@sil.org

Abstract

A generate and test algorithm is described which parses a surface form into one or more lexical entries using linearly ordered phonological rules. This algorithm avoids the exponential expansion of search space which a naive parsing algorithm would face by encoding into the form being parsed the ambiguities which arise during parsing. The algorithm has been implemented and tested on real language data, and its speed compares favorably with that of a KIMMO-type parser.

## 1. INTRODUCTION

A generate and test algorithm is described which uses linearly ordered phonological rules to parse a surface form into one or more underlying (lexical) forms.[*] Each step of the derivation may be rendered visible during both generation and test phases. The algorithm avoids an exponential expansion of search space during the generation phase by encoding the ambiguities which arise into the form being parsed. At the end of the generation phase, lexical lookup matches the ambiguous form against lexical entries. Because not all combinations of ambiguities in the parsed form are compatible, a test phase is used to filter forms found at lexical lookup. In this phase, the phonological rules are applied in forward order, and the derivations of any final forms which do not match the original input word are thrown out.

The algorithm has been implemented and tested on real language data; its speed is comparable to that of a KIMMO-type parser.

## 2. THE PROBLEM

Since the publication of *The Sound Pattern of English* (Chomsky and Halle 1968), most generative linguists have held that the phonological rules of natural languages are linearly ordered (Bromberger and Halle 1989). That is, when deriving a surface form from an underlying (lexical) form, the input of the N+1th rule is the output of the Nth rule.

While it is straightforward to derive a surface form from an underlying form with linearly ordered rules, complications arise in searching for the lexical form(s) from which a given surface form may be derived. One difficulty is that phonological rules are often neutralizing, so the result of "unapplying" such a rule during parsing is ambiguous. Consider the following simple rule:

[–continuant] --> [–voiced]

/ __ [–voiced]

Unapplication of this devoicing rule to a noncontinuant voiceless segment presents a dilemma: should the underlying segment be reconstructed as having been [+voiced], or was the segment originally [–voiced] (with the rule

---

having applied vacuously)? This dilemma arises under most theories with linearly ordered rules, whether segmental or autosegmental.

A second difficulty for parsing is that if rules apply in linear order, later rules can obscure the effects of earlier rules. In the example given, a later rule might alter the environment in which the devoicing rule had applied, e.g. by voicing a segment which served as the environment for the first rule.

This second problem arises in any theoretical framework which allows opaque rule orderings, that is, rule orders in which a later rule can opacify (obscure) the effects of earlier rules. Theories which disallow opaque rule orders (such as Natural Generative Phonology, see Hooper (1975)) have not enjoyed lasting popularity among linguists.

The implication of these two problems is that parsing would appear to require a bifurcation of the search space for each feature value assigned in the output of a phonological rule. For instance, consider the above devoicing rule, followed by a voicing rule which opacifies the first rule. Suppose we have a surface sequence of a voiceless noncontinuant segment followed by a voiced segment. In parsing this sequence, it would seem that we must explore several paths. If the surface voiced segment were also underlyingly voiced (vacuous application of the voicing rule), then there is no further choice; the surface voiceless noncontinuant could not have been devoiced by the devoicing rule. But if the surface voiced segment were underlyingly voiceless (nonvacuous application of the voicing rule), then the first rule might have applied, either vacuously or nonvacuously. Given that languages may have tens of phonological rules, and that each rule may alter multiple features, the search space becomes enormous.

Anderson (1988:5) summarizes the problem as follows:

> ...if the phonology of the language involves a non trivial amount of neutralization... it is necessary to calculate all of the possible combinations of alternatives allowed by various rules, which may be individually large when neutralizations are involved and whose product grows exponentially as the amount of significant rule interaction (ordering) increases.

> The combinatorial possibilities involved in undoing the phonology thus get out of hand rather quickly. Since the depth of ordering in a linguistically motivated description can easily approach 15-20, with many of the rules involved being many-ways ambiguous when regarded from the "wrong end," the approach of simply undoing the effects of the rules was soon seen to be quite impractical.

But in fact this expansion of search space can be avoided by the use of a generate-and-test algorithm, in which the ambiguity resulting from the unapplication of each rule is encoded into the form when the rule is unapplied. The resulting algorithm turns out to be tractable for the sorts of rules and rule ordering which arise in natural languages.

## 3. THE GENERATE-AND-TEST ALGORITHM

This section presents an algorithm for parsing with linearly ordered rules. The algorithm is efficient for the sorts of rule sets that have been proposed by generative phonologists for natural languages.

The algorithm is presented in general terms, abstracting away from implementational details where possible. Where a certain degree of concreteness is unavoidable—as in the definitions of the application or unapplication of a single rule—alternative forms of the algorithm are mentioned.

### 3.1 DEFINITIONS AND INITIAL ASSUMPTIONS

An *instantiated (phonetic) feature* is a feature-name plus an atomic feature value; an uninstantiated feature is merely the feature-name.

A *segment-specification* consists of a character representation of some segment (one or more characters, e.g. "k" or "ch"), plus a set of features, not all of which need be instantiated. An alphabet consists of a set of segment-specifications. A given language may employ more than one alphabet, distinguishing such as an input (surface) alphabet and a lexical (underlying) alphabet.

A *(phonetic) word* consists of a list of one or more segments, where each segment consists of a set of features. Input words (words to be parsed) and lexical words are usually represented instead in a character-based notation; the translation between this and a segment-based representation is defined below.

A *phonological rule* consists of an input (left-hand) side, an output (right-hand) side, a left environment, and a right environment. The *input* and *output* side each consist of a set of one or more instantiated features. (The extension to lists of sets, representing an input or output of more than a single segment, is straightforward. Rules in which the input or the output is empty, i.e. epenthesis or deletion rules, are discussed later.) The *environments* of a rule consist of a sequence of zero or more sets of instantiated features or optional sequences, together with a Boolean specification of whether the environment must begin (left environment) or end (right environment) at a word boundary. An *optional sequence* consists of a sequence of one or more sets of features, together with a minimum (MIN) and maximum (MAX) number of times the optional sequence may appear.

Finally, the *analysis target* of a rule is defined (for a rule with input and output of length one) as a set of features, which set consists of the features of the output, together with any non-contradictory features of the input. (In most rules, the features of the input and output are disjoint, so that the target consists of the union of the input and output features. Occasionally a rule will specify one value of a feature in the input, and a contrary value in the output. In that case, the analysis target takes the value of the feature in the output.)

A rule is said to be *self-opaquing* if it could be applied nonvacuously to a segment of its environments.[1] Such a rule must receive special treatment during analysis, because its application may have altered the word so that the output no longer meets the structural description of the rule.

The list of rules of a language is linearly ordered, and given in synthesis order. That is, the input of the first rule is a word from the lexicon, the input of the second rule is the output of the first rule, etc.; and the output of the last rule is a surface form.

### 3.2 TRANSLATION BETWEEN ALPHABETIC AND SEGMENTAL REPRESENTATIONS

A word in a phonetically based orthography (not, say, English orthography) may be translated into a segmental representation by the following algorithm:

---

[1]The precise formulation of "self-opaquing" for the purposes of the algorithm is somewhat more restrictive. Self-opaquing rules cause difficulty for parsing because such a rule may apply (nonvacuously) to some segment, while in the output the rule seems not to have applied to that segment because the environment for that segment has itself been altered by the rule so that it no longer meets the structural description: a self-counterbleeding rule. This can only happen if a segment of the environment meets the structural description of the rule, and the structural change of the rule assigns a value contrary to the value required in the environment. That is, a segment of the rule's environment is unifiable with the structural description of the rule but not with the structural change.

If the rule applies left-to-right iteratively, only the right environment is relevant, as only that environment can be altered after it has been used. Likewise, if the rule applies right-to-left iteratively, only the left environment is relevant. If the rule applies simultaneously, both environments are relevant.

Beginning at the left end of the word, replace the longest substring which corresponds to the character representation of some segment-specification in the appropriate alphabet, with its set of features.

Continue left to right, replacing substrings of the word with their features until the right end of the word is reached. If the process fails at any point (because no substring corresponds to a segment-specification), fail.

This translation algorithm is deterministic, and would give wrong results for a word like "mishap" (assuming "sh", "s" and "h" to be defined as segment-specifications). The algorithm could easily be made nondeterministic, with the proviso that each translation of an input word would be subjected to the remainder of the parsing algorithm. However, how multiple translations of lexical words would be treated is not so clear.

The translation between alphabetic and segmental representations could instead be done by a finite state transducer, with equivalent results.

### 3.3 UNAPPLICATION OF PHONOLOGICAL RULES

During the analysis phase of the algorithm, each rule is unapplied by uninstantiating in each segment which matches the rule in the correct environment, those features which the right-hand (output) side of the rule sets. For instance, if a rule assigns the value [-voiced] in its output, during parsing the value of the feature "voiced" in the segments affected by the rule becomes uninstantiated.

More specifically, given an input (surface) word in its segmental representation and a list of phonological rules, the rules may be unapplied to the word as follows.

(1) Reverse the list of rules to give a list in analysis order.

(2) Unapply the first rule of the list to the input word, using the algorithm below.

(3) Unapply each succeeding rule to the output of the previous rule.

The algorithm for the unapplication of a single rule in left-to-right iterative fashion (see Kenstowicz and Kisseberth 1979) is as follows; note that during analysis, a left-to-right iterative rule is applied right-to-left.

For each segment S beginning at the right end of the word:

If S is unifiable with the analysis target of the rule, and the left-hand environment of the rule matches against the word ending with the segment to the left of S, and the right-hand environment of the rule matches against the part of the word beginning with the segment to the right of S, then uninstantiate the features of S whose feature-names are contained in the output of the rule.

An environment sequence matches a subsequence of segments during analysis if:

For each member of the environment which is a set of features, that set unifies with the corresponding segment of the word; else (if the member is an optional sequence), the optional sequence matches against the corresponding sequence of segments between MIN and MAX number of times. If the environment must match at the margin of a word, then when the environment sequence is used up, the last segment matched must be the first segment of the word for the left environment, or the last segment for the right environment.

After a rule has been unapplied to a word, if the rule is self-opaquing and the unapplication was nonvacuous, the rule is unapplied again until its unapplication is vacuous.

The unapplication of a rule which applies right-to-left iteratively is the obvious transformation of the above algorithm.

The important point in the unapplication of a single rule to a form is the use of unification, so that a segment in the word matches a feature set in the rule even if the value of one or more relevant features in the segment has been uninstantiated by the unapplication of a previous rule. Matching against an uninstantiated feature thus represents an assumption, that the underlying value of that feature was correct. This assumption can only be validated during the synthesis phase, when a lexical entry from the lexicon will have become available.

The unapplication of a rule which applies simultaneously to its input may be performed by either left-to-right or right-to-left iterative unapplication, although the unapplication may need to be repeated if the rule is self-opaquing. To see why the self-opaquing test might be necessary, consider the following hypothetical rule:

[–sonorant] --> [+continuant]

/ __ [–continuant]

When applied simultaneously to the form *apkpa,* the result is *afxpa.* If the rule were unapplied to *afxpa* left-to-right iteratively, after the first pass we would have *af[x k]pa,* where the sequence *[x k]* is intended to represent a voiceless velar obstruent with an uninstantiated value for the feature [continuant] (hence ambiguous between the fricative *x* and the stop *k).* Only after a second pass would we get *a[f p][x k]pa.* (In this example the rule could have been unapplied right-to-left iteratively in a single pass, but a single right-to-left iterative application would have given the wrong result with the mirror image of the given rule.)

As an alternative to the above algorithm, the unapplication of a single rule could be performed by a Finite State Transducer (FST) (Johnson 1972, cf. also Kaplan and Kay, in press). It will be more convenient to compare the FST method with the above algorithm when we consider the application of a rule (as opposed to its unapplication).

### 3.4 LEXICAL LOOKUP

A word, some of whose segments may be partially instantiated, matches against a word in the lexicon if the features of each of its segments are unifiable with the corresponding segment of the lexical word. Lexical lookup consists of finding all such matches.

The unapplication of the phonological rules and the process of lexical lookup constitute the analysis phase of the algorithm.

### 3.5 APPLICATION OF PHONOLOGICAL RULES

As a result of the unapplication of rules to forms some of whose features may have been uninstantiated by earlier rules, some overgeneration may result, because a form taken from the lexicon may not have the value which was assumed during analysis. This overgeneration is filtered out by applying the rules in a synthesis phase. The degree of overgeneration is small, for reasons discussed in Maxwell (1991). The algorithm for applying rules during synthesis is straightforward:

> Given a lexical word and the list of rules, the first rule is applied to the lexical word, the second rule is applied to the output of the first, etc.

The application of a single rule in left-to-right iterative fashion is as follows:

> For each segment S beginning at the left end of the word:
>
> If S contains all the features of the left-hand side of the rule, and the left and right environments match parts of the word immediately to the left and right of S, then set the value of each feature in S whose name appears in the output of the rule to the value in that output.

An environment sequence matches during synthesis if:

> For each member of the environment which is a set of features, the corresponding segment of the word contains those same features; else (if the member is an optional sequence), the optional sequence matches against the corresponding segments of the word between MIN and MAX number of times. The condition on matching a word boundary is the same as during unapplication.

Right-to-left iterative application is again the obvious transformation of this algorithm. Simultaneous application may be modeled by first collecting the set of all segments which satisfy the structural description of the rule, and then applying the output of the rule to each segment in that set.

There is no need to check for possible reapplication of a rule during synthesis, as there was during analysis. This is because if the application of a rule creates new environments to which it might apply, those environments do not serve as further input for the rule apart from iteration or cyclic application. Directional iterative application is handled directly by the above algorithm, while nondirectional iterative application has generally been rejected by phonologists (cf. Johnson 1972: 35ff., and for a slightly different form of nondirectional iterative application, Kenstowicz and Kisseberth, 1979: 325). Cyclic application is not treated under the above algorithm, but would constitute only a restricted form of reapplication in which the application of a set of phonological rules would be sandwiched between each pair of cyclic morphological rules (as argued originally by Pesetsky 1979). If two or more cyclic morphological rules applied in a given word, the cyclic phonological rules would also apply at least twice. But each such application would be separated by the application of other rules, both phonological and morphological.

I will refer to this algorithm for applying a single rule as the Target-First Application Algorithm, or TFAA; it is analogous to the algorithm given earlier for unapplication of a rule.

As an alternative to the TFAA, each rule could instead be applied by an FST.

A disadvantage of application of a rule by the TFAA, compared with its application by FST, is that when checking the left-hand environment (assuming the rule applies left-to-right iteratively), the TFAA must retest segments it has already considered as possible target segments. In other words, the TFAA backs up through the form when checking the left-hand environment. Under those same circumstances, the FST need do no backing up when checking the left environment, as the applicability of the left environment is already determined when the FST arrives at a potential target. The distance the TFAA backs up can be considerable, in particular when the left environment (or the right environment, for a right-to-left iterative rule) has optional sequences (so that backtracking must be employed in case of failure to match the environment on the initial check), or when the word being parsed has "optional" segments. (Optional segments arise in analysis during the unapplication of deletion rules, as discussed later.)

Both the FST and the TFAA may test the same segments multiple times when the right-hand environment is nonempty (assuming left-to-right iterative application). For the FST, this will only happen if it made an incorrect choice. An example would be the rule:

[-continuant] --> [-voiced]

/ __ [-voiced]

when applied to the form *ba*. After the FST tests the target, it could attempt to apply the rule by assigning the feature [-voiced] to the *b* (changing it to *p)*. This would be incorrect, however, as the FST discovers when it processes the [+voiced] segment *a;* it must therefore back up, restore the [+voiced] value to the *b,* and move right to process the *a* again.

The TFAA, applying the same rule to the same form, would first notice the potential target *b*. Before altering the value of the feature [voiced], however, it would check the right environment: the segment *a*. Noticing that it does not satisfy the requirement that the right environment be [–voiced], it refrains from altering the feature [voiced] on the *b*. It then goes on to check whether the *a* constitutes a potential target.

However, the real question is not the worst case behavior, but the average case behavior; how many comparisons must be done for the average word with the average rule? Unfortunately, this is not a straightforward question. Examples are readily constructed in which the FST would do more comparisons than the TFAA. Given that in some cases the TFAA must back up through segments it has already considered while the FST need not, while in other cases the FST does more comparisons than the TFAA, I leave the question of average case behavior open. Note that similar considerations pertain to the behavior of the algorithm given earlier for the unapplication of rules.

A potential advantage of the TFAA over an FST implementation concerns the debugging of a single rule. When scanning a word for possible rule applications, people often search first for segments matching the input side of the rule, then check whether the left and right environments of potential targets also match. This is essentially the method employed in the TFAA. If a rule is at all complicated, trying to apply it as an FST instead becomes quite difficult for humans. By the same token, determining why a parser did or did not apply a rule to a certain segment of a form should be much easier if the parser presents a trace of its application in the same form that the human would do it. This is of course only an advantage of the TFAA if the user is actually tracing a given rule. Indeed the parser need not use the same algorithm to apply a rule when debugging is turned on as it uses when debugging is not turned on (although it is certainly easier on the writer of the parser if it does).

## 3.6 COMPARISON WITH INPUT FORM

Returning to the overall algorithm, specifically the test phase: the derivation of a word to which all the rules have been applied is correct if the derived word matches the original input word, that is, if each segment of the two words correspond. A segment corresponds if each of its features is identical.

During the test phase of the algorithm, a derived word may fail to match against the original (input) word under two circumstances: either one or more pairs of rules are opaquely ordered (see Maxwell 1991), or one or more rules are dependent on nonphonetic information, such as the location of a morpheme boundary or nonphonetic features. The resulting (potential) overgeneration is the reason for the test phase of the generate-and-test algorithm.

This completes the discussion of the generate-and-test algorithm for feature-changing rules. The next two sections discuss some refinements.

## 3.7 EPENTHESIS AND DELETION RULES

During analysis, a segment which has been inserted by an epenthesis rule[2] must be un-epenthesized, while segments which may have been deleted must be re-inserted. To avoid bifurcation of the search for each such segment, segments may be assigned an additional feature called "optional." All segments in the input word are marked [-optional]. When an epenthesis rule is unapplied (using an algorithm similar to that given above for feature-changing rules), the segments which might be epenthetic are marked as [+optional]. Similarly, a deletion rule may be unapplied by inserting a new segment with the set of features specified on the input side of the rule, and marking that segment as [+optional].

The unapplication of deletion rules must be further constrained to prevent infinite looping. To take a concrete example, consider the following consonant cluster simplification rule:

C --> 0 / C ____ C

If this rule is un-applied to a surface form with a two consonant cluster, the result will be an intermediate form having a three consonant cluster. But the rule is self-opaquing, in the sense that it can delete consonants which form part of the environment. Hence during analysis, it should be allowed to re-unapply to its own output. But if the rule is allowed to un-apply to the intermediate form produced by its first unapplication, namely a three consonant cluster, it can un-apply in two places to yield a five-consonant cluster, to which the rule can again be unapplied, ad infinitum.

---

[2]Pretheoretically, an epenthesis rule is a phonological rule which inserts a segment into a word. An example might be the insertion of *p* into *warm+th* to give *[warmpθ]*.

The best solution to this problem would be to use reasoning to determine the maximum number of contiguous consonants which could appear in the input to the rule. But this is by no means simple. It would be straightforward to determine the maximum number of consonants which could appear in underlying forms (based on the maximum number of consonants which appear in lexical entries and in affixes, assuming a morphological component), and in fact the lexicon itself is often used for this purpose in KIMMO-based systems. However, with linearly ordered rules the number of adjacent consonants could in principle be increased by the application of certain rules preceding the deletion rule, including rules epenthesizing consonants, rules deleting vowels, and rules changing vowels into consonants. Whether such rules in fact exist, or whether they exist but would be blocked by other principles from creating inputs to such a consonant cluster simplification rule is an area of research in phonology.

In the absence of a principled way of determining the maximum number of consonants that could appear in a cluster (or analogous limits on other deletion rules), an ad hoc limit may be placed on the application of deletion rules. One such limit is to unapply a deletion rule simultaneously, and only once (or only N times).

To take a concrete example, consider the input *abbabba,* where *a* is a vowel and *b* is a consonant. A single simultaneous unapplication of the above consonant cluster simplification rule would give *abCbabCba,* while two un-applications would give *abCCCbabCCCa,* where the first and third *C*s in each cluster result from the second unapplication. Limiting the un-application of deletion rules in this way is ad hoc, but probably sufficient for practical purposes.

The presence of [+optional] segments arising from the unapplication of epenthesis and deletion rules slightly complicates the algorithm given earlier for rule unapplication, in that such segments may optionally be passed over when checking rule environments.

During synthesis, epenthesis rules are straightforwardly applied by inserting a segment with the features of the output of the rule, while deletion rules are applied by simply deleting the relevant segments.

### 3.8 NONPHONETIC FEATURES, BOUNDARY MARKERS, ALPHA FEATURES ETC.

Nonphonetic (diacritic) features and obligatory boundary markers in rules may simply be ignored during analysis, leading to some overgeneration. In (manually) checking a number of such rules against large dictionaries, overgeneration appears to be surprisingly small, in fact virtually nil.

Alpha variable features (commonly used in assimilation rules) may be modeled by the use of variables which become instantiated to the value of features in the appropriate segments, so that checking for a match during analysis is a matter of unification. During synthesis, a variable in the output of a rule results in the features of the corresponding segment of the word being set to the value to which the variable becomes instantiated in some other part of the rule.

### 4. AN IMPLEMENTATION OF THE ALGORITHM

The generate-and-test algorithm has been implemented, as a parser which uses phonological rules of classical generative phonology, resembling those of Chomsky and Halle (1968) and much related work. (A sample rule is shown in the appendix.) I call the parser "Hermit Crab." There is provision for feature-changing rules (including alpha variable rules), epenthesis rules, and deletion rules. Disjunctive rule ordering may be modeled, as well as simultaneous or directional iterative application. The environments of rules may incorporate optional sequences (such as $(CV)_1^2$).

PC-KIMMO, an implementation of two-level phonology (Antworth 1990) was used to provide a comparison between parsing with linearly ordered generative phonological rules, and with two-level rules. Both PC-KIMMO and Hermit Crab run under MS-DOS.

PC-KIMMO comes with example analyses of the phonologies of several languages, including Hebrew, Turkish, Japanese, and Finnish, each analysis containing from 16 to 27 two-level rules. The PC-KIMMO analyses were converted into analyses using linearly ordered generative rules, which were equivalent in the sense that they derived the surface forms from the same underlying forms. In most cases the linearly ordered rules were simpler than the two-level rules, in part because rule ordering rendered redundant some of the constraints necessitated by the two-level formalism. The number of rules for each language was reduced to between 7 and 11, as some two-level rules (such as default rules) are unneeded in a generative analysis, while others collapse into disjunctively ordered rule sets. For instance, PC-KIMMO has six rules for vowel harmony in Turkish: two for backness harmony in low vowels (one to make a low vowel [+back] in the appropriate environment, and one to make it [–back] in the opposite environment), and four rules for backness and rounding harmony in nonlow vowels. These collapse into two generative rules: one for backness harmony, which affects all vowels, and uses an alpha variable for the two possible values of the feature back; and one rule for rounding harmony, which affects nonlow vowels, again using an alpha variable for the two possible values of the feature *round*.

Because the focus here is on phonological parsing, rather than morphological parsing, the morphological rules given in PC-KIMMO's sample analyses were ignored, and fully affixed forms were used for underlying forms, e.g.:

<lex_entry shape "oda+sH"

... gloss "room+POSS">

In a sample of several hundred words, PC-KIMMO was about three times faster than the parser using linearly ordered rules. This difference is not large, and indeed may be attributed in part to the different programming languages used (PC-KIMMO is written in C, while the parser implementing the generate-and-test algorithm is written in Prolog and C). The ratio of 3:1 is approximately constant among the four grammars, and independent of word length, indicating that the results should scale.

5. CONCLUSION

The algorithm as described and implemented models segmental phonology. An extension to multiple strata of rules, as in lexical phonology, is trivial, and has also been implemented. Allowing cyclic application of rules is also simple, although it has not been implemented yet (because most phonologists since Pesetsky 1979 have interpreted cyclic phonology as the interleaving of phonological rules and morphological rules, and morphological rules have not yet been implemented).

The algorithm could be extended to autosegmental models of phonology by reinterpreting e.g. feature spreading rules as feature assignment rules with alpha variables during the analysis phase, and reverting to the standard interpretation of autosegmental rules during synthesis. For instance, an autosegmental rule spreading the place of articulation features of an obstruent onto a preceding nasal consonant can be modeled during analysis by the following rule:

$$\begin{bmatrix} +\text{cons} \\ +\text{nasal} \end{bmatrix} \rightarrow \begin{bmatrix} \alpha \text{ high} \\ \beta \text{ back} \\ \gamma \text{ coronal} \end{bmatrix} / \_\_ \begin{bmatrix} -\text{continuent} \\ \alpha \text{ high} \\ \beta \text{ back} \\ \gamma \text{ coronal} \end{bmatrix}$$

The modeling of autosegmental phonology has not been implemented, although the use of alpha variables has.

In summary, and contrary to many earlier claims, it need not be computationally expensive to parse surface forms into their underlying forms using linearly ordered rules. Furthermore, unlike rule compilers (Kaplan and Kaye in press), the use of a rule interpreter simplifies grammar debugging, as the input and output of each rule can be studied (a sample trace is shown in the appendix).

## 7. APPENDIX: A SAMPLE PARSING RUN

This appendix presents excerpts of the input to and output from the phonological parser. Some information in the original has been omitted here for simplicity. The language being parsed is Japanese. The data and rules were adapted from one of the sample data sets provided with PC-KIMMO (see Antworth 1990). The rule notation is an internal format, not necessarily intended for the end user. Structures are enclosed in "< >". Added comments are shown in italics.

*We first load the rules. The rule shown here ([+voc –round] --> 0 / [+voc] + __) deletes a non-round vowel immediately after another vowel, but note the obligatory morpheme boundary ("+"):*

```
(load_morpher_rule
<prule   rname vowel_deletion
         p_lhs <p_lhs pseq ((+ voc –round))>
         p_rhs <p_rhs pseq ()>
         left_environ <ptemp pseq ((+ voc) "+")>>)
```

*...remaining rules are not shown here.*

*Turn tracing on ('T') for lexical lookup:*

```
(trace_lexical_lookup T)
```

*Also turn tracing on for the vowel deletion rule, both during analysis (the first 'T') and synthesis (the second 'T'):*

```
(trace_morpher_rule (T T vowel_deletion))
```

*Finally, parse the Japanese word "neta":*

```
(morph_and_lookup_word "neta")
```

*The parser's trace output follows. First the trace repeats the input:*

<trace shape "neta"
    continuations (

*The "continuation" list is a list of paths followed by the parser from the current position. The parser proceeds by unapplying the rules. The trace shows the unapplication of the traced rule, "vowel_deletion". The form which is input to this rule is the sequence of segments resulting from the unapplication of shallower rules, and corresponds to the regular expression "n([r y])et([r y])a", where "[r y]" signifies a segment ambiguous between "r" and "y" which the parser has undeleted; the parentheses indicate that it is optional (since perhaps the parser shouldn't have undeleted it). Unapplication of the traced deletion rule results in the insertion of more optional segments in two places, whose features correspond to the set of vowels /i e a/:*

<rule_unapp rname vowel_deletion
    input <lex_entry shape "n([r y])et([r y])a"... >
    output <lex_entry shape "n([r y])e([i e a])t([r y])a([i e a])"... >
    continuations (

*Continuing, the parser unapplies various other rules (not shown here, since they aren't being traced); after all the rules have been unapplied, it does lexical lookup:*

    <lex_lookup virtual
        <lex_entry shape "n([r y])e([i e a])t([r y])a([i e a])"...>

*The above "virtual" (i.e. created during analysis) lexical entry corresponds to a "real" one found in the lexicon:*

        continuations (
        <succ_lookup real <lex_entry shape "ne+ta" gloss "(sleep)+PAST"... >
            continuations (

*The parser continues from this lexical lookup by applying the rules to the form in the synthesis phase; again, only one application is shown, the traced one. The rule does not apply, since its structural description is not met; hence the output form is the same as the input form:*

        <rule_app rname vowel_deletion
            input <lex_entry shape "ne+ta" gloss "(sleep)+PAST" ...>
            output <lex_entry shape "ne+ta" gloss "(sleep)+PAST"...>
            continuations (

*After applying the remaining rules, a surface form results whose phonetic shape is identical to the input form:*

        <lex_entry shape "neta" gloss "(sleep)+PAST"...>)>)>

*Now we return to continuations from lexical lookup. There is a second "real" lexical entry in the user's lexicon which corresponds to the "virtual" lexical entry the parser created:*

    <succ_lookup real
        <lex_entry shape "ne+itai" gloss "(sleep)+VOL"...>
        continuations (

*Again, the continuation from the lexical item consists of the application of the rules; only the traced rule's application is shown. The deletion rule applies to the first pair of adjacent vowels, turning /ei/ into /e/, but does not apply to the second pair of vowels /ai/, although during analysis it was unapplied in both places. The reason is that the rule requires a morpheme boundary to appear between the vowels, and there is only a morpheme boundary between the first pair of vowels. (During analysis, the position of morpheme boundaries is unknown, hence they are impossible to check for.) This analysis will be filtered*

*out; it is an example of overgeneration due to a rule whose application is governed by morpheme boundaries:*

<rule_app rname vowel_deletion
    input <lex_entry shape "ne+itai" gloss "(sleep)+VOL"...>
    output <lex_entry shape "ne+tai" gloss "(sleep)+VOL"...>

*The continuation list from this rule application is empty, because the output does not match the input word:*

    continuations ( )

>)>)>)>)>

*This completes the trace. Last comes the parser's output, the single analysis of the word, depicted as a lexical entry at the surface level. This would have been output even if tracing had not been turned on:*

(word_analysis
    <lex_entry shape "neta" gloss "(sleep)+PAST"... > )